\documentclass[preprint,12pt,authoryear,nopreprintline]{elsarticle}

\usepackage{amsmath}
\usepackage{amssymb}

\usepackage{graphicx}
\usepackage{dcolumn}
\usepackage{bm}
\usepackage[hidelinks]{hyperref}

\usepackage{siunitx} 
\usepackage{mathrsfs}

\usepackage{comment}
\usepackage{booktabs}
\usepackage{pgfplots}
\pgfplotsset{compat=1.18}
\usepackage{scalerel}
\usepackage{tikz}

\hypersetup{pdfauthor=Steven Goldenberg}

\journal{Machine Learning with Applications}

\begin{document}

\title{Distance Preserving Machine Learning for Uncertainty Aware Accelerator Capacitance Predictions}

\author[1]{Steven Goldenberg\corref{cor1}}
\ead{sgolden@jlab.org}
\author[1]{Malachi Schram}
\ead{schram@jlab.org}
\author[1]{Kishansingh Rajput}
\ead{kishan@jlab.org}
\author[1]{Thomas Britton}
\ead{tbritton@jlab.org}

\author[2]{Chris Pappas}
\ead{pappasgc@ornl.gov}
\author[2]{Dan Lu}
\ead{lud1@ornl.gov}
\author[2]{Jared Walden}
\ead{waldenjd@ornl.gov}
\author[3]{Majdi I. Radaideh}
\ead{radaideh@umich.edu}
\author[2]{Sarah Cousineau}
\ead{scousine@ornl.gov}
\author[4]{Sudarshan Harave}
\ead{haraves@slac.stanford.edu}
\affiliation[1]{organization={Thomas Jefferson National Accelerator Facility}, city={Newport News}, postcode={VA 23606}, country={USA}}
\affiliation[2]{organization={Oak Ridge National Laboratory}, city={Oak Ridge}, postcode={TN 37830}, country={USA}}
\affiliation[3]{organization={Department of Nuclear Engineering and Radiological Sciences, The University of Michigan}, city={Ann Arbor},
postcode={MI 48109}, country={USA}}
\affiliation[4]{organization={SLAC National Accelerator Laboratory}, city={Menlo Park}, postcode={CA 94025}, country={USA}}

\cortext[cor1]{Corresponding author}

\date{\today}

\begin{abstract}
Providing accurate uncertainty estimations is essential for producing reliable machine learning models, especially in safety-critical applications such as accelerator systems. 
Gaussian process models are generally regarded as the gold standard method for this task, but they can struggle with large, high-dimensional datasets. Combining deep neural networks with Gaussian process approximation techniques have shown promising results, but dimensionality reduction through standard deep neural network layers is not guaranteed to maintain the distance information necessary for Gaussian process models. 
We build on previous work by comparing the use of the singular value decomposition against a spectral-normalized dense layer as a feature extractor for a deep neural Gaussian process approximation model and apply it to a capacitance prediction problem for the High Voltage Converter Modulators in the Oak Ridge Spallation Neutron Source. 
Our model shows improved distance preservation and predicts in-distribution capacitance values with less than 1\% error.
\end{abstract}


\maketitle

\section{Introduction}

Machine learning (ML) and deep neural networks (DNNs) provide extraordinary accuracy for prediction of complex systems when paired with large datasets like those produced by the Oak Ridge National Laboratory (ORNL) Spallation Neutron Source (SNS). 
Models trained with these vast amounts of available accelerator data can improve accelerator availability by forecasting anomalies and pending failures in complex systems. 
However, as ML and DNNs become increasingly relevant to these complex and critical applications, it is important to provide models that are both reliable and trustworthy. 
In order to do this, ML models need to provide a well calibrated uncertainty estimations to avoid catastrophic system failures if poor predictions arise from previously unseen input data. 
These poor predictions are particularly likely when the data-driven ML model is applied on out-of-distribution (OOD) data.

\begin{figure}
    \centering
    \includegraphics[width=0.8\linewidth]{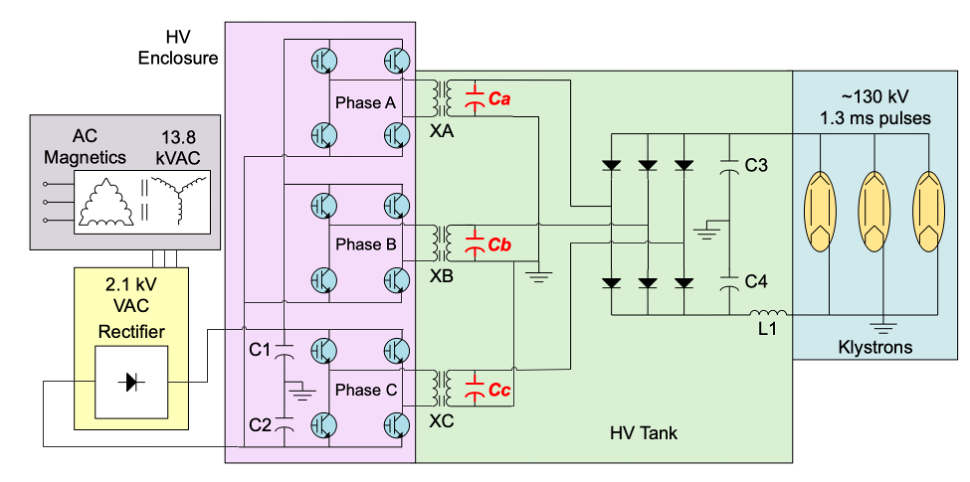}
    \caption{Simplified schematic of an HVCM system. The three capacitors that require capacitance predictions (Ca, Cb, and Cc) are highlighted in red.}
    \label{fig:hvcm_schematic}
\end{figure}

This work focuses on using ML with uncertainty quantification (UQ) to predict the capacitor degradation within the High-Voltage Converter Modulators (HVCMs) in the SNS. 
A simplified schematic of an HVCM \citep{Reass2003DesignSNSHVCM} and klystron load is shown in Figure \ref{fig:hvcm_schematic}. 
The resonant capacitors, Ca, Cb, and Cc at the secondaries of the pulse transformers have recently caused significant downtime due to failure. For brevity, we call these capacitors A, B and C respectively in the rest of this paper. 
These capacitors have historically been a film/foil type of construction, but they are now being replaced by metallized film capacitors which show a loss of capacitance as they age before eventual failure \citep{Flicker2013LiftimeCapacitor}. 

As capacitance measurements require nearly a full week to collect, we only have access to a very limited set of labeled real data. Instead, we leverage a LTSpice simulation of the HVCM to learn a relationship between capacitance and waveforms available in the real data from existing sensors \citep{LTspice-website}. There were several deficiencies with earlier simulations when compared to real HVCM data due to differences in switch timing between simulated and real, as well as differences in the voltage settings \citep{radaideh:napac2022-wepa38}. The LTSpice model now includes tuning of the IGBT switches timing to match that of real HVCM waveform data.

We propose using a DNN for producing accurate predictions of the capacitance because they are known to scale well to the large and high-dimensional datasets available. 
Recent work on neural Gaussian processes (GP) has shown significant promise for a single forward pass model when paired with approximation techniques like random Fourier features (RFF) or inducing point methods \citep{vanAmersfoort2020DUQ, vanAmersfoort2021DUE, rahimi2007RFF, liu2020SNGP}. 
Additionally, Gaussian process approximations (GPA) have been used to provide UQ for other accelerator systems at the SNS and Fermilab National Accelerator Laboratory \citep{Blokland2022UQML-ErrantBeamSNS, Schram2023UQML-FNAL}.

Deep Neural Gaussian Process Approximation (DNGPA) methods also benefit from explicit input distance awareness when care is taken to preserve pairwise distances from the input layer to the latent space used to build the Gaussian process covariance matrix which is used predicting OOD uncertainty. 
With high-dimensional data, feature reduction is often required, but doing so may result in drastic changes in the distances between samples in the reduced space, which potentially results in unreliable uncertainty estimations. Ideally, our DNN appropriately preserves meaningful distances from the input space, however previous work by \citep{liu2020SNGP} mainly focuses on distance preservation within dimensionally identical layers. 
This work focuses on maintaining sample distances while performing dimensionality reduction within the context of DNGPA methods. 

To do this, we evaluate the use of the singular value decomposition (SVD) for performing the dimensionality reduction to maintain sample distances and improve uncertainty estimates when the model is applied to both in-distribution (ID) and OOD data. This work is a novel extension of \citep{liu2020SNGP} and further emphasizes the importance of distance awareness for uncertainty quantification and OOD prediction. 
We compare our model to three other methods for producing uncertainty estimates in DNNs and show significant improvements in predictive power and highly accurate ID uncertainty estimates. 
Our model allows for nearly real-time predictions of the three capacitance levels in a non-invasive way, reduce system downtime, and avoid the cost of additional sensors. 
Trends in this predicted capacitance data could also inform the performance of preventative maintenance including replacing worn components before failure.

\section{Data Preparation}

\begin{figure}
    \centering
    \includegraphics{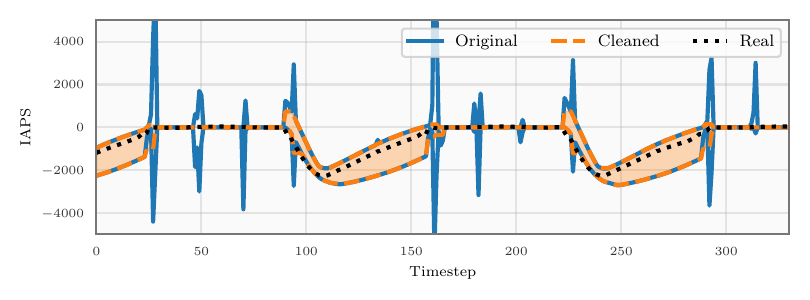}
    \caption{Range over all simulation data (min/max) for the IAPS current waveform before and after cleaning with a LULU filter showing a significant reduction in excursion artifacts.}
    \label{fig:ICleaning}
\end{figure}

Given the challenges in gathering training data for HVCMs throughout its capacitors' lifetime, we relied on synthetic data sets gathered from LTSpice simulations \citep{radaideh:napac2022-wepa38}.
These simulations contain a variety of artifacts; discontinuities born from how the simulation converges to it's final solution.  These artifacts, especially those that have magnitudes greater than the underlying waveforms, profoundly impact the data normalization and the ability of the ML method to ``learn'' the correlations between the waveforms themselves and the values of capacitance present in the HVCM.

To combat this problem, the traces representing the currents in the HVCM (the only traces containing excursion artifacts) were cleaned.  
Cleaning was performed by applying a LULU filter \citep{rohwer1989LULU}, which provides a fast and idempotent algorithm to effectively remove impulsive errors in the simulation data. 
This kind of filter may alter accurate data points by a minimal amount, but drastically improves all excursions and provides a stable range for training. 
As an example, Figure \ref{fig:ICleaning} shows a portion of an example trace from the real system as well as the range of simulation samples (min/max) before and after cleaning. 

The total dataset contains 1792 samples with 7 waveforms containing 5261 timesteps each. 
These seven waveforms include ``V\_out'' and six current waveforms: ``IAPS'', ``IAP'', ``IBPS'', ``IBP'', ``ICPS'', ``ICP''. 
The six current waveforms are measured from the positive buses of the three H-bridge phases (``IAP'', ``IBP'', ``ICP'') with a star (``S'') and non-star waveform determined by the direction of power flow in each phase (controlled by pairs of transistors).
These simulations come from two sets of capacitor values. 
In 100 picofarad (pF) increments, the first set ranges from 2500 to 2800 pF for OOD testing (64 samples), while the other ranges from 2900 to 4000 pF and is randomly split for in-distribution training and testing (1382 and 346 samples respectively). 

\begin{figure}
        \centering
        \includegraphics{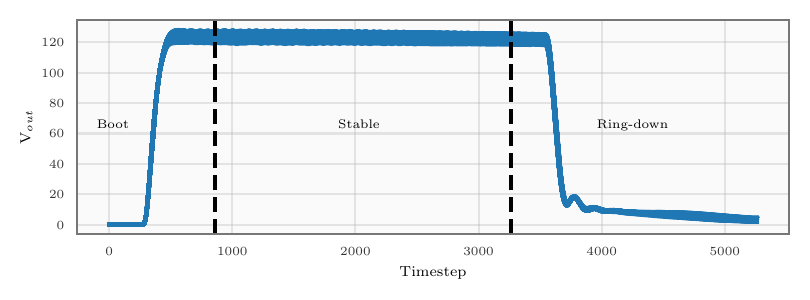}
        \caption{A sample V$_{out}$ waveform illustrating the 3 regions: boot, stable, ring-down.}
        \label{fig:3regions}
\end{figure}

Three distinct regions were identified in the simulated waveforms: ``boot'', ``stable'', and ``ring-down'' (see Figure \ref{fig:3regions}). 
The boot region extends from index zero to approximately index 860. 
The stable region comprises the bulk of the waveform from index 860 to index 3260. 
The remaining timesteps comprise the ring-down region. 
Only the final 1000 timesteps of the stable region is considered in this analysis as it is the region with the best agreement with real data and reduces the quantity of highly repetitive portions of the stable waveform.

\section{Methods and Techniques}

The review by Abdar et al.\citep{abdar2021UQreview} gives a comprehensive background of currently available methods for UQ in deep learning models. 
While there are many available options, our work compares methods such as Monte Carlo (MC) Dropout \citep{gal2016dropoutUQ}, Deep Quantile Regression (DQR) \citep{koenker2005QRBook} and Spectral Normalized Gaussian Processes (SNGP) \citep{liu2020SNGP}. 
Our new approach modifies an SNGP model by altering the feature reduction step necessary for high-dimensional data. 
We provide more detailed descriptions of these models in the following subsections.

Our work does not consider ensemble methods for producing UQ, however we use the ensemble method to estimate the variability and robustness of each solution.
Specifically, this provides a mean and standard deviation for our accuracy and uncertainty calibration metrics in Section \ref{sec:Results}. 

\subsection{Deep Quantile Regression}

Unlike standard linear regression which approximates the conditional mean over the training data, quantile regression attempts to estimate quantiles of a response variable. 
Therefore, it can be more robust to outliers as it uses the conditional median as a prediction. 
Additionally, because the outputs are conditioned on the desired quantile, we can estimate multiple quantiles at once to obtain uncertainty estimations. 
Deep quantile regression (DQR) extends this idea by applying quantile regression techniques to deep learning models to learn more complex functions. 

Given a conditional quantile $\tau$, and input feature vector $x$, we can define the conditional quantile function:
\[Q_y(\tau | x) = G_\tau(x , w)\]
where $G_{\tau}(x,w)$ is a non-linear function described by a DNN. For each desired quantile, the loss function used for regression is given by
\[\mathcal{L}(y,\hat{y}) = \max(\tau(y - \hat{y}), (\tau - 1)(y - \hat{y}))\]
where $y$ and $\hat{y}$ are the label and prediction of the model respectively. 

In this paper, we define our desired quantiles as $\tau = [0.159, 0.5, 0.841]$. 
The median provides a robust prediction while the first and last quantiles match the expected proportions for one standard deviation from the mean in a normal Gaussian distribution. 
To obtain a single uncertainty estimation, we take the average of $G_{0.5}(x, w) - G_{0.159}(x, w)$ and $G_{0.841}(x, w) - G_{0.5}(x, w)$. 
We note that the differences between these quantiles may be large as they are not guaranteed to be symmetric around the median. Nonetheless, these choices allow for direct comparisons with the other models in this paper. 

\subsection{Bayesian Neural Networks}
MC-Dropout \citep{gal2016dropoutUQ} has become a popular model for estimating prediction uncertainty in DNNs due to its ease of implementation and benefits as a model regularizer which prevents over-fitting of the network. 
Additionally, it can be viewed as a slightly cheaper approximation to a Bayesian model, however its relationship to a true Bayesian model has recently been called into question \citep{folgoc2021MCDnotBayesian}. 

MC-Dropout estimates the uncertainty by computing the standard deviation from a set of inferences where each inference differs by the location of randomly removed nodes between layers (by setting values to 0). 
Since uncertainty predictions for BNNs come from multiple inferences, as opposed to a single model output, this model is significantly slower for both training and final inferences when compared to the other models we implemented. 

Our implementation of MC-Dropout includes a trainable parameter for the dropout probability, which allows the model to fit the uncertainty estimations using the log-loss:
\begin{equation}\label{eq:log-loss}
    \mathcal{L}_{dropout} = \frac{1}{N} \sum_{i=1}^N \frac{\|y - \hat{y} \|^2}{2\sigma(x)^2}  + \frac{\log(\sigma(x)^2)}{2}
\end{equation}
 This loss function is based on the negative log-likelihood and appropriately balances the standard deviation with the squared residual \citep{kendall2017UQLogLoss, williams2006GPBook}. 

\subsection{GP Methods}
\label{subsec:GP_Methods}

GP methods are often considered ideal for uncertainty estimation as they become more uncertain as test input samples move away from the training data. 
However, the cost of a standard GP regression model on $n$ training samples is $O(n^3)$ which is infeasible for large datasets. 
Approximation methods like RFFs or inducing point methods can significantly improve this by bounding computation through limiting the size of the covariance matrix used for UQ. 

In order to produce uncertainty estimates, GPs calculate the covariance of the conditional joint Gaussian distribution given by
\[cov(f^*) = K(X_*, X_*) - K(X_*, X)[K(X, X) + \sigma^2_n I]^{-1} K(X, X^*), \]
where $X$ and $X_*$ are the training and testing input data, $\sigma_n^2$ is a noise variance term, and $K$ represents a chosen kernel function. 
The kernel function is very flexible and may have many parameters in order to change the covariance calculations. 
Our model calculates an RFF approximation to the Gaussian radial basis function (RBF) kernel with a trainable parameter ($\lambda$) that controls the width of distribution. 
The formula for the standard Gaussian RBF is given below:
\begin{equation}\label{eq:Gaussian_RBF}
    K(x_1,x_2) = e^{\frac{-\|x_1 - x_2\|^2}{2\lambda^2}}. 
\end{equation}
The noise variance $\sigma_n$ is also a trainable parameter within our model. 
Like our BNN model, we train these two parameters using the loss from Equation \ref{eq:log-loss}. 
It is important to note that, unlike a typical GPA model, the kernel that produces the uncertainty estimations is not used to compute predictions. 
Instead, the predictions are computed through a dense neural network layer without bias that takes the RFFs as an input. 

\subsection{Distance Preservation Techniques}

GP rely on the distance between inputs to accurately represent uncertainties due to their inclusion in the kernel function \ref{eq:Gaussian_RBF}. 
If a DNN is used to reduce the dimensionality of the input prior to a GP layer, the latent representation may distort these distances, causing worse performance. 
Ideally, we would like the DNN to maintain bi-Lipschitz constraints such that:
\begin{equation} \label{eq:bilipschitz}
    L_1 \|x_1 - x_2\| \leq \|h(x_1) - h(x_2)\| \leq L_2 \|x_1 - x_2\|. 
\end{equation}
Here $h(x_1)$ is the latent representation of input sample $x_1$.

\subsubsection{Spectral Normalization}

\cite{liu2020SNGP} demonstrated the issues caused by lack of distance preservation and attempted to solve them by limiting the spectral norm of dense layers inside of residual networks. 
This works because of the skip connections that enforce the latent representation to be $h(x)=x+f(x)$. 
Since spectral normalization enforces a bound on the spectral norm of $f(x)$, these layers avoid changing distances too drastically. In fact, \cite{liu2020SNGP} presents a proof that given a maximum spectral norm, $\alpha \leq 1$, the bi-Lipschitz constraints in Equation \ref{eq:bilipschitz} can be maintained for $l$ layers with $L_1=(1- \alpha)^{l-1},\; L_2=(1+ \alpha)^{l-1}$. 

One aspect not explored in their work was distance preservation when reducing the dimension of the input space, as the residual network can no longer directly use the input for the skip connections. Instead, the output of a residual layer will be $h(x)=g(x)+f(x)$ where we require a $g(x)$ that provides dimensionality reduction while maintaining distances. 
\cite{liu2020SNGP} references several other works that describe methods of preserving approximate isometry after significant dimensionality reduction, however, these methods were not used in their work. 
The work presented in this paper explores the effects of two different methods for dimensionality reduction: a spectral-normalized dense layer and the singular value decomposition (SVD).

\subsubsection{Singular Value Decomposition}

The singular value decomposition along with other orthogonal projection methods like principal component analysis and the eigenvalue decomposition, are well-studied algorithms for producing a rank-reduced representation of datasets. 
Specifically, the SVD produces a decomposition of the data matrix, $X \in \mathbb{R}^{d\times n}$ such that $X=U\Sigma V^T$ where $U$ and $V$ are orthonormal bases and $\Sigma$ is a diagonal matrix containing the singular values of $X$. 
This decomposition can be truncated to produce a weight matrix with a given output dimension by using only the first $k$ columns of $V$. 

Unlike a spectral normalized dense layer, these methods produce orthonormal weight vectors from the columns of $V$ which guarantees both the spectral norm and the smallest singular value of the weight matrix are 1. 
While this is a stricter requirement than spectral normalization, we can maintain model expressiveness through the subsequent residual network. 

Additionally, this constraint significantly improves our distance preservation in many cases by optimizing the norm preservation for the training data. 
Specifically, given a sample $x$ (i.e. row of $X$), we have 
\[ \|x\| - \sqrt{\sum_{j=k+1}^n \sigma_j^2} \leq \|xW \| \leq \|x\|. \] 
The proof of this inequality is provided in \ref{appendix:svd_proof}. 
The lower bound on $\|xW\|$ can provide guarantees for sample norm preservation as the last $n-k$ singular values are often very small for approximately low-rank data matrices. 
Additionally, this bound is very conservative as it assumes the last $n-k$ entries of the row of $U$ corresponding to $x$ are all 1, which cannot be true due to the orthonormal columns of $U$. 
A more reasonable assumption would be that each entry is a $\mathcal{N}(0,1\sqrt{d})$ Gaussian random variable which would scale the expectation of the summation in the bound by $d^{-0.25}$. 

In order to compute a truncated SVD, we require multiplications with the entire training dataset on a set of k+b vectors (where b is some relatively small oversample) which requires O(nd(k+b)) time. 
With very large and high-dimensional data, this may seem impractical. 
However, there are algorithms like incremental SVD \citep{baker2012incrementalSVD} that compute the SVD in a batched way that matches the way most DNNs are trained. 
Assuming batches larger than $k$, we expect this method to produce similar results to the ones we present in this paper.  

\subsection{Model Parameters}

\begin{figure}
    \centering
    \includegraphics[width=0.95\linewidth]{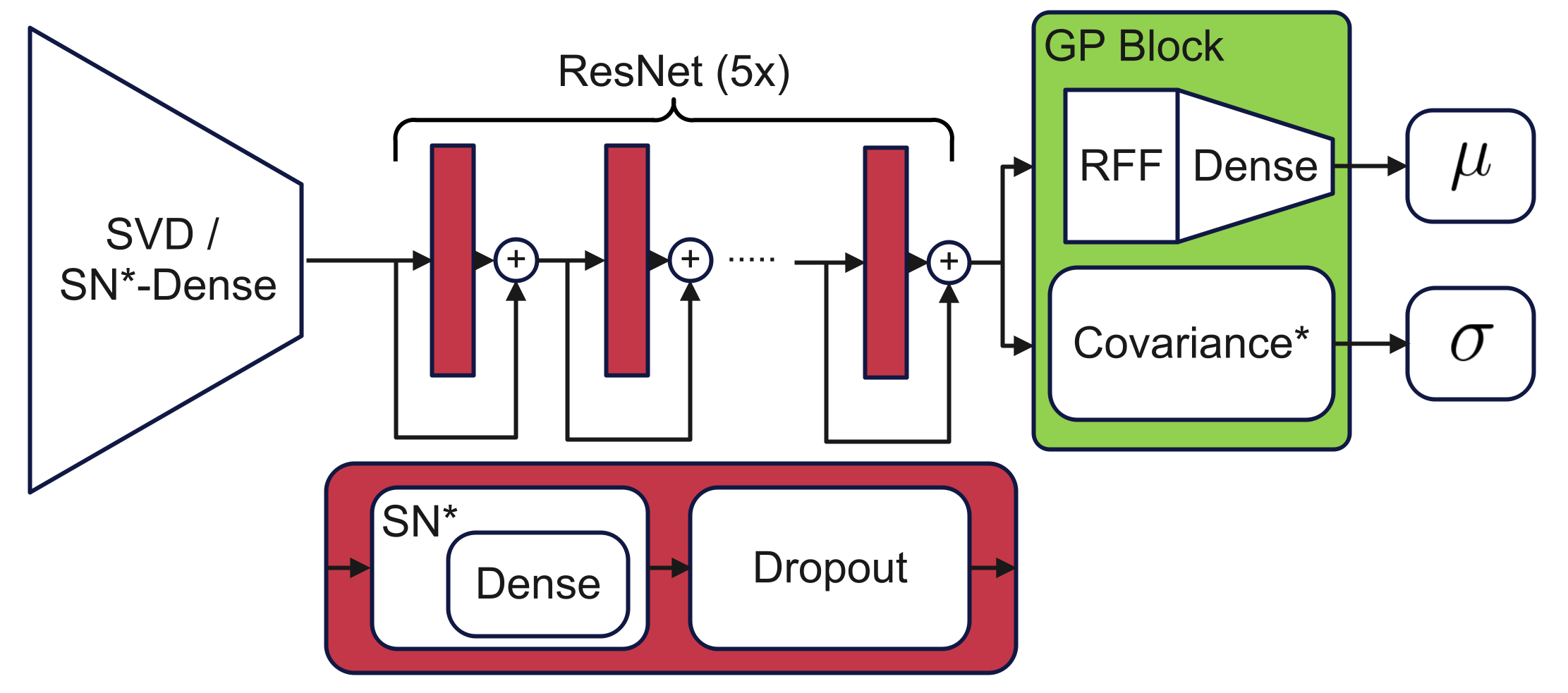}
    \caption{Model architecture diagram. SVD-DNGPA is the only model that uses the SVD feature extraction as the first “layer”. Layers marked with * are only used in the DNGPA models in order to better preserve distances and provide UQ.}
    \label{fig:model_diagram}
\end{figure}

\begin{table}[h]
    \centering
    \begin{tabular}{ll} \toprule
         Model Parameter & Value \\ \midrule
         Optimizer & ``adam'' \\
         Latent Dimension & 64 \\
         Random Fourier Features & 128 \\
         ResNet Activations & ``relu'' \\
         Dropout Rate & 10\% \\
         ResNet SN Constant* & 0.8 \\
         Input SN Constant$^\dagger$ & 1.2 \\ \bottomrule
    \end{tabular}
    \caption{Parameters used for our ML models. Values marked with * are only used for our two DNGPA based models, and values marked with $^\dagger$ are only used by the base DNGPA model.}
    \label{tab:model_params}
\end{table}

All models used in this work include a feature extraction layer on the flattened dataset, a 5-layer residual network (ResNet) with Dropout, an RFF layer, and a Dense output layer for predictions as seen in Figure \ref{fig:model_diagram}. 
Even though the RFF layer is not required by the DQR and BNN models to provide UQ, we chose to include it to maintain a consistent forward path for each model.
For information about model parameters, see Table \ref{tab:model_params}. 

However, our models differ in a few key areas.
The initial feature reduction layer for the SVD-DNGPA model utilizes the SVD while all other models tested used a Dense layer. 
Moreover, we apply the spectral normalization constraint only for Dense layers used by the DNGPA models, since distance preservation is not required for the DQR and BNN models to produce accurate uncertainties.
Lastly, our BNN model uses a trainable value for the Dropout layer to optimize its uncertainty estimates, while all other models use a consistent 10\% dropout rate.

\section{Results}
\label{sec:Results}

In this section, we test our new model against a more standard DNGPA model, a BNN approximation (MC-Dropout), and DQR. 
Each model is trained 15 times with different random initial seeds to create an ensemble of models. 
This ensemble provides a mean and standard deviation for each metric we report. 
Additionally, we quantify their performance for both ID and OOD predictions and uncertainties. 
For OOD testing, we introduce a shift in the output labels. 
The best models should be able to generalize to these OOD samples and should increase their uncertainty estimations as the sample labels move further away from the training basis.

\subsection{GP Model Distance Preservation}

\begin{figure}
    \centering
    \includegraphics{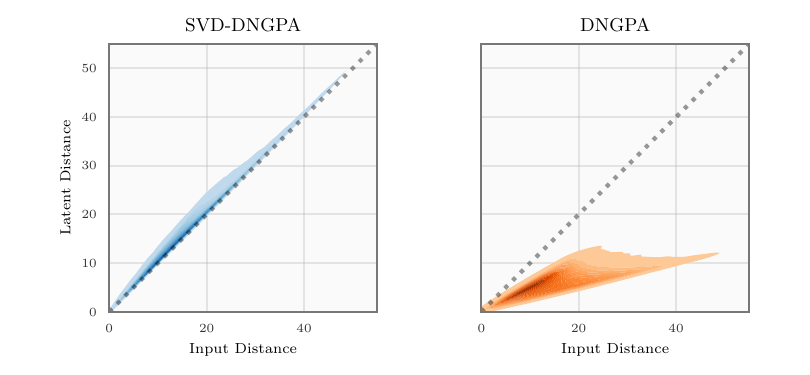}
    \caption{Correlation between input and hidden Euclidean distances for the SVD-DNGPA and DNGPA models. Plots are created with a kernel density estimation over a random 10\% sample of the distances to avoid building a full kernel for 1382 training data samples.}
    \label{fig:dist_corr}
\end{figure}

For GP models, as uncertainties are derived from a kernel based on the input to the GP layer, it is critical to maintain a strong correlation between the GP input in the latent space and the original input. 
As an initial test for the bound we derived for the SVD norm preservation, we computed the first 64 singular values and compared them with the Frobenius norm of the data matrix. 
Specifically, on our training data and $k = 64$, we compute:
\[ \sqrt{\sum_{j = k+1}^n \sigma_j^2} = \sqrt{\|X\|_F^2 - \sum_{j=1}^k \sigma_j^2} = 11.97.\]
This can be compared with the norm of our original samples which is bounded by $\|x\| > 39.10$. 
The true amount of norm degradation caused by the SVD alone is even smaller than the scaled bound with $\max(\|x\| - \|xW\|) \approx 0.015$. 
This is further reinforced in Figure \ref{fig:dist_corr}, which shows how using the SVD in our model significantly improves the correlations between input distances and distances in the latent space compared to a more standard spectral-normalized dense layer. As desired, both the spread of the correlation plot and the deviations from the optimal correlations (dotted line) are reduced. 

\subsection{In-Distribution Results}

\setlength{\tabcolsep}{0.5em}
\begin{table*}[]
    \centering
    \small
    \begin{tabular}{lll} \toprule
         Model & R$^2$ & RMSE (pF) \\ \midrule
SVD-DNGPA & \num{9.998e-01} $\pm$ \num{1.379e-05} & \num{5.150e+00} $\pm$ \num{1.614e-01}  \\
DNGPA     & \num{9.998e-01} $\pm$ \num{2.945e-05} & \num{5.457e+00} $\pm$ \num{3.229e-01}  \\
BNN       & \num{9.994e-01} $\pm$ \num{1.815e-04} & \num{8.672e+00} $\pm$ \num{1.207e+00}  \\
DQR       & \num{9.997e-01} $\pm$ \num{7.161e-05} & \num{5.902e+00} $\pm$ \num{6.871e-01}  \\ \bottomrule
    \end{tabular}
    \caption{Mean and standard deviation of our accuracy results on an ID test set with 15 random initializations of each model.}
    \label{tab:ID-pred-results}
\end{table*}

\setlength{\tabcolsep}{0.5em}
\begin{table*}[]
    \centering
    \small
    \begin{tabular}{lll} \toprule
         Model & RMSCE & MACE \\ \midrule
SVD-DNGPA & \num{2.284e-02} $\pm$ \num{1.331e-02} & \num{1.930e-02} $\pm$ \num{1.200e-02} \\
DNGPA     & \num{8.163e-02} $\pm$ \num{6.804e-03} & \num{7.200e-02} $\pm$ \num{6.852e-03} \\
BNN       & \num{1.843e-01} $\pm$ \num{3.670e-02} & \num{1.650e-01} $\pm$ \num{3.295e-02} \\
DQR       & \num{2.429e-02} $\pm$ \num{4.789e-03} & \num{1.984e-02} $\pm$ \num{5.135e-03} \\ \bottomrule
    \end{tabular}
    \caption{Mean and standard deviation of our uncertainty calibration results on an ID test set with 15 random initializations of each model.}
    \label{tab:ID-uq-results}
\end{table*}

Deep learning models are specifically trained to obtain accurate predictions and uncertainty estimations on ID evaluation samples as they are generally consistent with the training data. 
To compare the accuracy of our models, we calculate the coefficient of determination (R$^2$) and the Root Mean Squared Error (RMSE). Given a perfect model, these values should be 1 and 0 respectively. 
For our uncertainty estimations, we calculate the Root Mean Squared Calibration Error (RMSCE) and the Mean Absolute Calibration Error (MACE) using the Uncertainty Toolbox (Chung et al. 2021), where values closer to zero indicate a well-calibrated model. 
In order to ensure our results are robust, we trained an ensemble of 15 models for each method with varying random seeds to produce a mean and standard deviation for each metric.  
Tables \ref{tab:ID-pred-results} and \ref{tab:ID-uq-results} show the results of these tests.

\begin{figure*}
    \centering
    \includegraphics{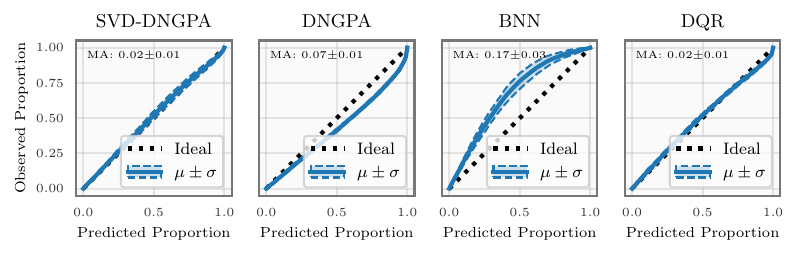}
    \caption{Miscalibration area (MA) plots for the four model types averaged over 15 trials. Plots with a larger shaded region indicate inconsistent performance caused by the random initializations. The mean and standard deviation of the MA are given in the top left for each model.}
    \label{fig:id_miscal_plots}
\end{figure*}

While all models perform well on our ID test set, SVD-DNGPA has slightly better average performance for every metric. 
All models have an RMSE less than 10pF which is significantly less than the 10\% manufacturer tolerances for these capacitors ($\approx 300$pF). 
We note that while DQR is the second-best method in terms of the average calibration performance, it has a smaller standard deviation indicating that this method may produce marginally more robust uncertainty estimations. 
To further illustrate the performance of these methods for uncertainty estimations, we provide Figure \ref{fig:id_miscal_plots}, which shows a modified miscalibration area plot with the mean and standard deviation over 15 trials for each model. 
SVD-DNGPA provides very accurate uncertainties with very little variation from one trial to the next. 
This result is a significant improvement on the standard DNGPA model and rivals DQR for our in-distribution testing.

To be thorough, we also tested our models by changing the size of the latent space to see if BNN would improve or if other trends were present. 
We found that smaller models (e.g. latent size of 32) often exhibited more variability in the test results although DQR seemed less affected by the change. 
We believe this may be due to the loss function used by the BNN and DNGPA models, which can explode if the standard deviation approximation is too small. 
On the other hand, we saw very little difference between a latent space size of 64 and 128 and therefore chose the smaller model for efficiency. 

\subsection{Out-of-Distribution Results}
While it is always preferable to train on the full range of possible inputs/outputs, practical limitations often prevent full exploration. 
For example, obtaining real data where outputs represent an unstable system may create a safety risk. 
Furthermore, in a dynamic system, distribution drift and anomalous behaviors can arise which may not be easily known prior to training the ML model. 
In theory, the best models should be able to generalize well to OOD data and provide higher uncertainty when moving away from the training basis. 

\begin{table*}[]
    \centering
    \small
\begin{tabular}{ l l @{\ $\pm$\ }l  r @{\ $\pm$\ }l } \toprule
  \multicolumn{1}{l}{Model} & \multicolumn{2}{l}{R$^2$} & \multicolumn{2}{l}{RMSE} \\ \midrule
SVD-DNGPA & \quad\!\num{8.396e-01} & \num{1.396e-01}  & \num{4.047e+01} & \num{1.918e+01} \\
DNGPA     & \quad\!\num{4.985e-01} & \num{4.592e-01}  & \num{6.891e+01} & \num{3.899e+01} \\
BNN       & \num{-1.210e+00} & \num{2.391e+00} & \num{1.461e+02} & \num{7.926e+01} \\
DQR       & \num{-7.882e-01} & \num{1.389e+00} & \num{1.385e+02} & \num{5.633e+01} \\
  \hline
\end{tabular}
    \caption{Mean and standard deviation of our accuracy results on an OOD test set with 15 random initializations of each model.}
    \label{tab:OoD_pred_results}
\end{table*}

\begin{table*}[]
    \centering
    \small
    \begin{tabular}{lll} \toprule
         Model & RMSCE & MACE \\ \midrule
SVD-DNGPA & \num{3.814e-01} $\pm$ \num{1.413e-01} & \num{3.337e-01} $\pm$ \num{1.214e-01} \\
DNGPA     & \num{3.985e-01} $\pm$ \num{1.397e-01} & \num{3.470e-01} $\pm$ \num{1.188e-01} \\
BNN       & \num{2.995e-01} $\pm$ \num{1.684e-01} & \num{2.604e-01} $\pm$ \num{1.472e-01} \\
DQR       & \num{3.292e-01} $\pm$ \num{1.408e-01} & \num{2.909e-01} $\pm$ \num{1.241e-01} \\
    \end{tabular}
    \caption{Mean and standard deviation of our uncertainty calibration results on an OOD test set with 15 random initializations of each model.}
    \label{tab:OoD_uq_results}
\end{table*}

To test our models for this behavior, we measured the same statistics as our ID test, but on a set of samples where all three capacitors had capacitance values of 2800pF and below. 
We don’t expect any of our models to provide accurate uncertainties, as our models have not been tuned for this OOD data. 
However, it is still very desirable to have reasonable predictions and uncertainties that grow consistently based on their distance from the original training data. 
Tables \ref{tab:OoD_pred_results} and \ref{tab:OoD_uq_results} show the results of our models over 15 random initializations. 
We see that all models perform significantly worse than the training data for predictions as the RMSE scores increased by nearly an order of magnitude and the standard deviation for our results increased by nearly two orders. 
However, SVD-DNGPA still provides reasonable results with an R$^2$ value greater than 0.8 and an RMSE that was two to four times smaller than all of the other models we tested. 
For completeness, we also provide the RMSCE and MACE scores, but all models perform within a factor of approximately 1.5 and have large standard deviations relative to the score, which is expected given the OOD nature of the test set.

\begin{figure}
    \centering
    \includegraphics{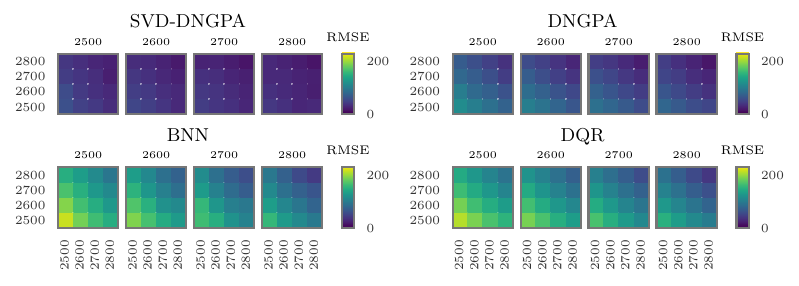}
    \caption{RMSE over 15 trials on the OOD test set. The 4 grids for each model represent the A capacitance label, while the B and C capacitance labels are given as the $y$ and $x$ axes respectively. As capacitance values get smaller (further from our training data), we see increased RMSE for all models.}
    \label{fig:OoD_RMSE}
\end{figure}

To further analyze the models, we plot the RMSE in Figure \ref{fig:OoD_RMSE} for each triplet of capacitance values in our test set. 
As expected, the RMSE degrades as we move towards smaller output labels (moving right to left, top to bottom). 
Additionally, the RMSE is significantly smaller for the two DNGPA models, suggesting that these models are much better at generalizing to OOD data for this application.

\begin{figure*}
    \centering
    \includegraphics{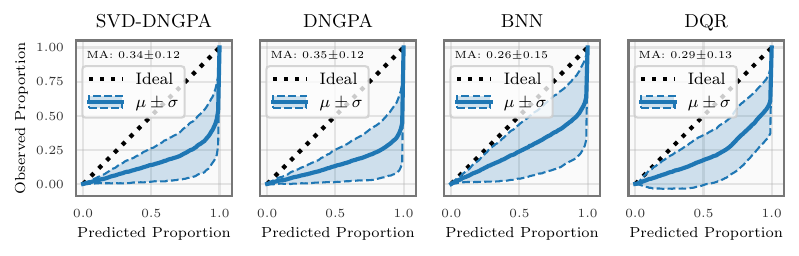}
    \caption{Miscalibration area (MA) plots for the four model types averaged over 15 trials on the OOD test set. Plots with a larger shaded region indicate inconsistent performance caused by the random initializations. The mean and standard deviation of the MA are given in the top left for each model.}
    \label{fig:OoD_miscal_plots}
\end{figure*}

We assess whether the models are well calibrated for OOD values in Figure \ref{fig:OoD_miscal_plots}. 
All models perform equally poorly and show no significant statistical difference, which coincides with the large RMSCE and MACE values reported in Table \ref{tab:OoD_uq_results}. 
We note that the BNN model consistently has higher variance for in both ID and OOD testing, suggesting that it may be less reliable overall.
While these results are somewhat disappointing, it is important to recognize that none of these models have been tuned to appropriately estimate uncertainty for OOD data. 
It may be possible to improve results for the GP methods by choosing a better prior, but the BNN and DQR models cannot be improved without further training data.

\section{Discussion and Future Work}
\label{sec:Discussion}

While our results are very promising especially for ID samples, it is important to discuss potential short-comings as well as interesting new ideas generated by our work. 
For example, the results from this paper are derived from a simulation that while highly representative is ultimately synthetic. The real data does not always map directly to our simulated data, particularly when the SNS configuration is changed. We hope to obtain real labeled data from a single-phase low voltage system similar to the three-phase high voltage SNS HVCM, which we can use to train a model in the future. 

We also hope to obtain more labeled data from the real three-phase HVCMs to further validate the accuracy of the models we trained on simulations. 
We currently have two sets of data for pulses that were gathered relatively soon after capacitor changes (sufficiently close so as to expect no significant degradation has taken place). 
As two labels is insufficient to draw any strong conclusions (especially with many potentially confounding factors), we leave this validation for future work.

Additionally, although our DNGPA models produce accurate predictions and uncertainties, there are opportunities to improve their performance through kernel parameter optimization. 
For future work, we would like to further investigate alternative kernel options beyond RBF and whether our learned RBF parameters are similar to those learned through an exact GP regression technique. 

Next, our HVCM dataset is amenable to a simple linear dimensionality reduction like the SVD, but many other problems lie in a non-linear space with low rank. 
Additionally, Euclidean distance metrics may not capture the true distance of points that lie on a non-linear hyperplane. 
We would like to investigate further how these problems could be solved with other approximate isometric maps like Isomap \citep{tenenbaum2000ISOMAP} or other algorithms as mentioned in \citep{perrault2012MetricLearning}. 
It may also be interesting to investigate how to preserve distances over physics-constrained manifolds. 

Lastly, we would like to explore whether RFF layers can improve other large models by reducing the model size. 
While this layer is required for our GP models, we found that removing this layer from our non-GP models reduced performance by an order of magnitude. 
We believe the difference in performance stems from the RFF layer introducing a significantly larger amount of non-linearity compared to a a standard dense layer with a non-linear activation. 

\section{Conclusion}
In systems like the HVCMs at ORNL, providing accurate predictions and uncertainty estimations is a requirement in order to avoid system failure, detect anomalous conditions, and provide information for appropriate prevention measures. 
By enforcing stronger distance preservation techniques within a DNGPA model, we have shown improvements in the robustness of both in-distribution uncertainty estimations and OOD predictions when compared with current state-of-the-art model architectures for uncertainty quantification. 
In addition, we have illustrated the direct impact of the SVD with regards to distance preservation compared to a spectral-normalized dense layer and proved bounds on the SVD norm preservation. 
Our new model architecture handles high-dimensional data well by avoiding feature collapse and only requires a single inference step. We believe these ideas and their future extensions are crucial for providing distance preservation in neural networks with dimensionality reduction. 

\section{Acknowledgments}
The authors acknowledge the help from David Brown in evaluating Operations requirements, and Frank Liu for his assistance on the Machine Learning techniques.

This manuscript has been authored by UT-Battelle, LLC, under contract DE-AC05-00OR22725 with the US Department of Energy (DOE). The Jefferson Science Associates (JSA) operates the Thomas Jefferson National Accelerator Facility for the U.S. Department of Energy under Contract No. DE-AC05-06OR23177. This research used resources at the Spallation Neutron Source, a DOE Office of Science User Facility operated by the Oak Ridge National Laboratory. The US government retains and the publisher, by accepting the article for publication, acknowledges that the US government retains a nonexclusive, paid-up, irrevocable, worldwide license to publish or reproduce the published form of this manuscript, or allow others to do so, for US government purposes. DOE will provide public access to these results of federally sponsored research in accordance with the DOE Public Access Plan (http://energy.gov/downloads/doe-public-access-plan).

\bibliographystyle{elsarticle-harv} 

\bibliography{refs}

\appendix

\section{Proof of SVD Norm Preservation}
\label{appendix:svd_proof}

For any given sample (i.e. row of $X$) $x$ and the SVD $X = U \Sigma V^T$, we have $x = u \Sigma V^T$ where $u$ is the corresponding row of $U$. Note that this notation for $u$ is somewhat contrary to the standard where $u$ denotes the orthonormal columns of $U$. Given a split of $V = \begin{bmatrix} V_1 & V_2 \end{bmatrix}$ where $V_1 = W$ are the weights used for dimensionality reduction, applying these weights produces:
\begin{equation}
\begin{split}
    x W &= u \Sigma V^T W \\
          &= u \Sigma \begin{bmatrix} I \\ 0 \end{bmatrix} \\
          &= \begin{bmatrix} u_1 \sigma_1 & \cdots & u_k \sigma_k \end{bmatrix}
\end{split}
\end{equation}
Similarly, $x V_2 = \begin{bmatrix} u_{k+1} \sigma_{k+1} & \cdots & u_n \sigma_n \end{bmatrix}$. Therefore, $xV = \begin{bmatrix} x W & 0\end{bmatrix} + \begin{bmatrix} 0 & x V_2\end{bmatrix}$. To simplify notation, we denote these zero extended versions of $x W$ and $x V_2$ as $z_1$ and $z_2$ respectively and note that their norms are equivalent (i.e. $\|z_1\| = \|x W\|$). Additionally, no value in $u$ is greater than 1, otherwise the singular vectors would not have unit norm. Assuming all values in u are 1 gives us $\|z_2\| \leq \sqrt{\sum_{j=k+1}^n \sigma_j^2}$.  which we can combine with the triangle inequality to obtain the following inequality:
\begin{equation}
\begin{split}
        \|x\| &= \|xV\| = \|z_1 + z_2\| \leq \|z_1\| + \|z_2\| \\
        &\leq \|xW\| + \sqrt{\sum_{j=k+1}^n \sigma_j^2} . \\
\end{split}
\end{equation}

After rearranging terms, we obtain our desired relation:
$\|x\| - \sqrt{\sum_{j=k+1}^n \sigma_j^2} \leq \|xW\| $. Also, since W is orthonormal, it is clear that $\|xW\| \leq \|x\|$. 
To strengthen this inequality further, we can assume that each entry in $U$ is $\mathcal{N}(0, 1/\sqrt{d}),$ which would provide vectors with an expected unit length. Doing this gives the following equation which can reduce the expected sample norm reduction by a factor based on the number of data samples.

\begin{equation}
    \begin{split}
        E\left[\sqrt{\sum_{j=k+1}^n \sigma_j^2 u_j^2}\right] &= \sqrt{\frac{1}{\sqrt{d}}\sum_{j=k+1}^n \sigma_j^2} \\
        &= d^{-0.25} \sqrt{\sum_{j=k+1}^n \sigma_j^2},
    \end{split}
\end{equation}
This equation suggests an interesting balance when adding new data to X. Essentially, increasing the number of samples improves our norm preservation, but new data may add variance to previously under-represented directions which would increase the truncated singular values.

\end{document}